\documentclass[10pt,twocolumn,letterpaper]{article}
\pdfoutput=1
\usepackage{cvpr}
\usepackage{times}
\usepackage{epsfig}
\usepackage{graphicx}
\usepackage{amsmath}
\usepackage{amssymb}

\usepackage{array}
\usepackage{url}

\usepackage[pagebackref=true,breaklinks=true,letterpaper=true,colorlinks,bookmarks=false]{hyperref}

\cvprfinalcopy 

\def\cvprPaperID{18} 

\ifcvprfinal\fi
\begin{document}

\title{Visualizing and Describing Fine-grained Categories as Textures}

\author{Tsung-Yu Lin \quad Mikayla Timm \quad Chenyun Wu \quad Subhransu Maji\\
University of Massachusetts, Amherst\\
{\tt\small \{tsungyulin,mtimm,chenyun,smaji\}@cs.umass.edu}
}

\maketitle


We analyze how categories from recent FGVC
challenges~\cite{fgvc5,fgvc6} can be described by their \emph{textural content}.
The motivation is that subtle differences between species of birds
or butterflies can often be described in terms of the texture associated
with them and that several top-performing networks are inspired by texture-based
representations.
These representations are characterized by orderless pooling of second-order
filter activations such as in bilinear
CNNs~\cite{lin2018bilinear} and the winner of the iNaturalist 2018
challenge~\cite{Li_2018_CVPR}. 

\begin{figure}
\centering
\setlength{\tabcolsep}{2pt}
\begin{tabular}{ccc}
\includegraphics[width=0.3\linewidth]{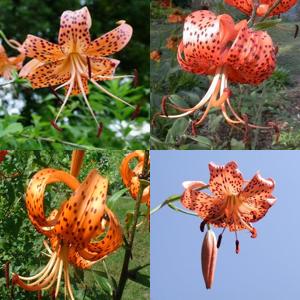}
& 
\includegraphics[width=0.3\linewidth]{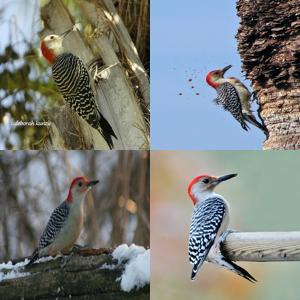}
&
\includegraphics[width=0.3\linewidth]{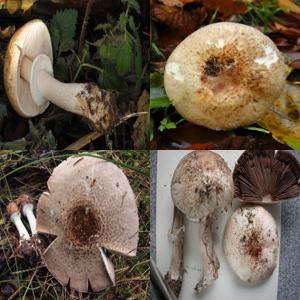}
\\
\includegraphics[width=0.3\linewidth]{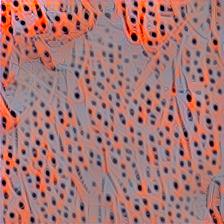}
&
\includegraphics[width=0.3\linewidth]{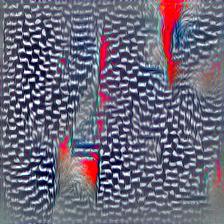}
&
\includegraphics[width=0.3\linewidth]{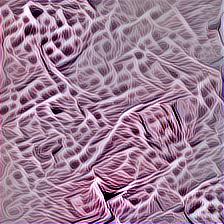}
\\
\includegraphics[trim={1cm, 0, 0, 0},
  clip,width=0.3\linewidth]{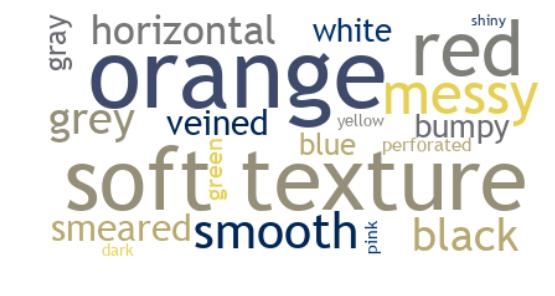}
& 
\includegraphics[trim={1cm, 0, 0, 0}, clip,
  width=0.3\linewidth]{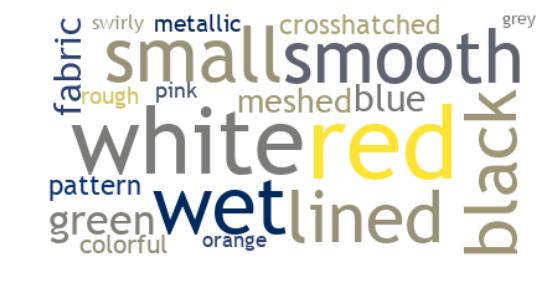}
&
\includegraphics[trim={1cm, 0, 0, 0}, clip,
  width=0.3\linewidth]{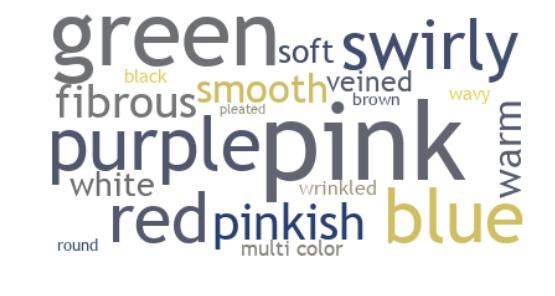}
\end{tabular}
\caption{\label{fig:splash}\emph{Tiger Lily} (left), \emph{Red Bellied
    Woodpecker} (middle) and \emph{Boletus Reticulatus} (right)
  categories visualized as their training images (top row), maximal
  texture images (middle row) and texture attributes
  (bottom row). The size of each phrase in the cloud reflects its
  likelihood of being associated with the maximal texture.}
\vspace{-0.1in}
\end{figure}

Concretely, for each category we (i) visualize the
``maximal images'' by obtaining inputs $\mathbf{x}$ that maximize the
probability of the particular class according to a texture-based
deep network $C_\theta(\mathbf{x})$, and (ii) automatically describe
the maximal images using a set of texture attributes.
We use $C_\theta$ as a multi-layer bilinear CNN as described
in our prior work on visualizing deep texture
representations~\cite{lin2016visualizing}.
The models for texture captioning were trained on our ongoing efforts
on collecting a dataset of describable textures building on the DTD dataset\cite{cimpoi14describing}.
As seen in Figure~\ref{fig:splash}, these visualizations indicate what aspects of the texture is most
discriminative for each category while the descriptions provide a
language-based explanation of the same.

\vspace{-0.2in}
\paragraph{Visualizing categories as maximal textures.}
We visualize the categories from Caltech-UCSD birds~\cite{WahCUB_200_2011}, Oxford
flowers~\cite{Nilsback08}, FGVC flowers~\cite{fgvcflowers},
FGVC fungi~\cite{fgvcfungus} and FGVC butterflies and moths~\cite{fgvcbutterlies} datasets.
Following the approach of~\cite{lin2018bilinear} we extract the covariance matrix
followed by signed square-root and $\ell_2$ normalization from
\textit{relu\{2\_2,3\_3,4\_3, 5\_3\}} layers of VGG-16 network~\cite{simonyan2014very} and
train a softmax layer to predict class labels. 
We train the model on the standard training split for birds and Oxford flowers and
randomly select 100 images from the 200 categories with the most
images for FGVC fungi, flowers, and butterflies.

\begin{figure*}[t]
\begin{center}
\setlength{\tabcolsep}{2pt}
\begin{tabular}{ccc|ccc}
\multicolumn{3}{c|}{\textbf{Caltech-UCSD Birds}}  & \multicolumn{3}{c}{\textbf{Oxford Flowers}} \\
\includegraphics[width=0.15\linewidth]{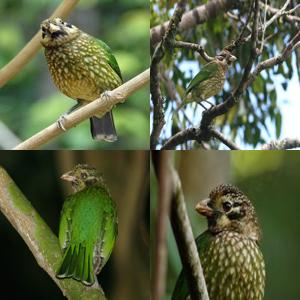} &
\includegraphics[width=0.15\linewidth]{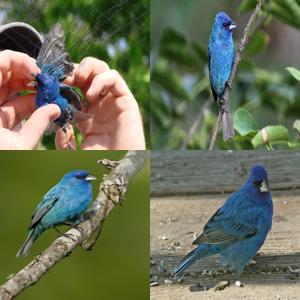} &
\includegraphics[width=0.15\linewidth]{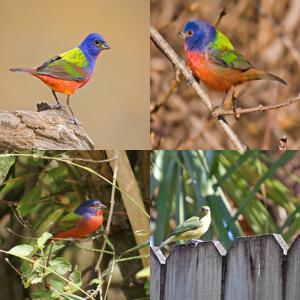} &
\includegraphics[width=0.15\linewidth]{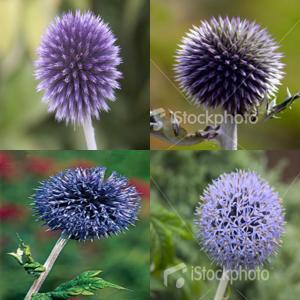} &
\includegraphics[width=0.15\linewidth]{fig/montage_images/oxford_flowers/class_6.jpg} &
\includegraphics[width=0.15\linewidth]{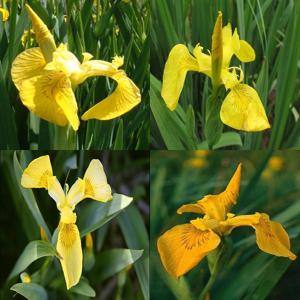} \\
\includegraphics[width=0.15\linewidth]{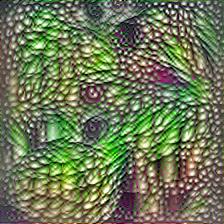} &
\includegraphics[width=0.15\linewidth]{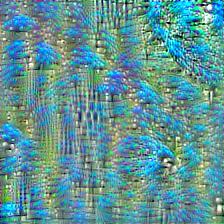} &
\includegraphics[width=0.15\linewidth]{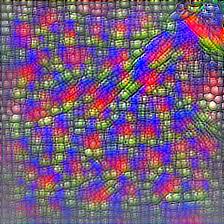} &
\includegraphics[width=0.15\linewidth]{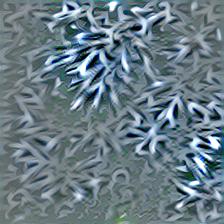} &
\includegraphics[width=0.15\linewidth]{fig/texture_images/oxford_flowers/class_6.jpg} &
\includegraphics[width=0.15\linewidth]{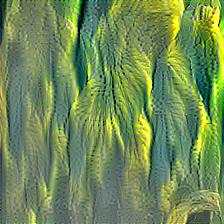} \\
\includegraphics[trim={1cm, 0, 0, 0}, clip, width=0.15\linewidth]{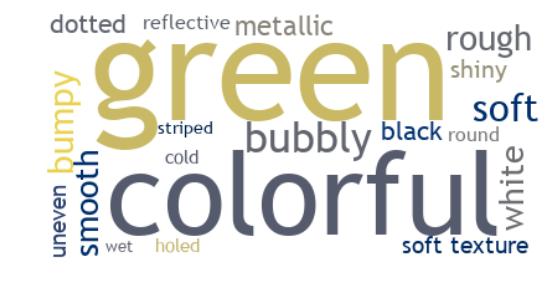} &
\includegraphics[trim={1cm, 0, 0, 0}, clip, width=0.15\linewidth]{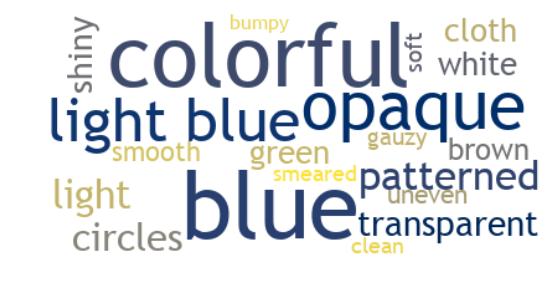} &
\includegraphics[trim={1cm, 0, 0, 0}, clip, width=0.15\linewidth]{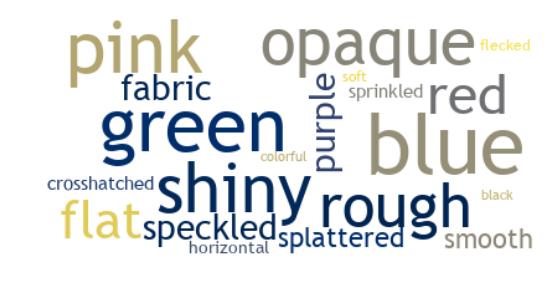} &
\includegraphics[trim={1cm, 0, 0, 0}, clip, width=0.15\linewidth]{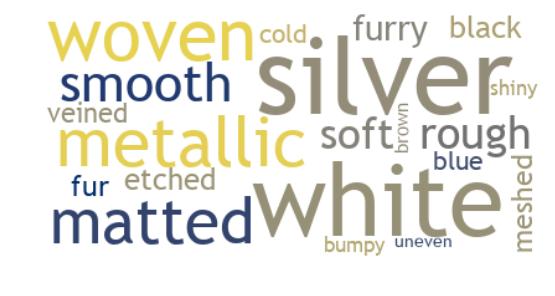} &
\includegraphics[trim={1cm, 0, 0, 0}, clip, width=0.15\linewidth]{fig/word_images/oxford_flowers/class_6.jpg} &
\includegraphics[trim={1cm, 0, 0, 0}, clip, width=0.15\linewidth]{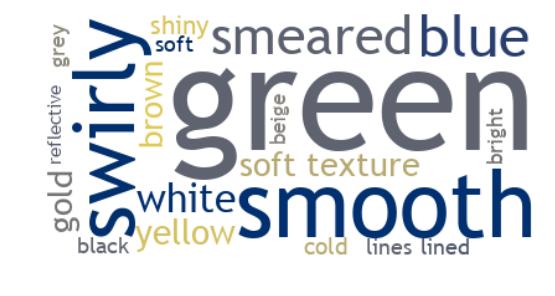} \\
\includegraphics[width=0.15\linewidth]{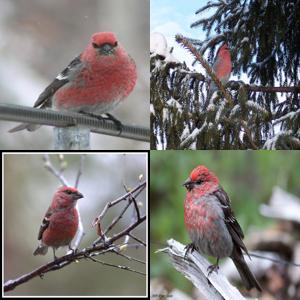} &
\includegraphics[width=0.15\linewidth]{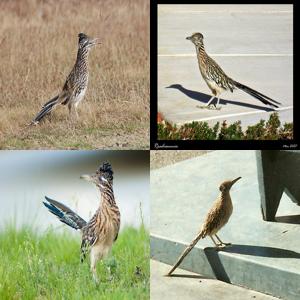} &
\includegraphics[width=0.15\linewidth]{fig/montage_images/cub/189_Red_bellied_Woodpecker.jpg} &
\includegraphics[width=0.15\linewidth]{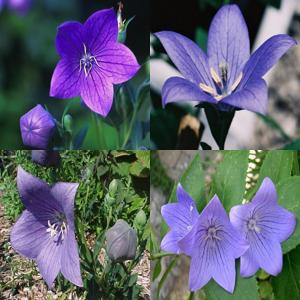} &
\includegraphics[width=0.15\linewidth]{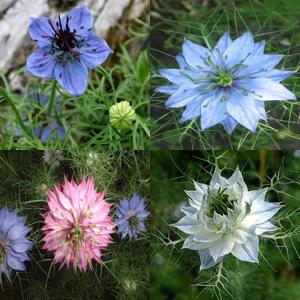} &
\includegraphics[width=0.15\linewidth]{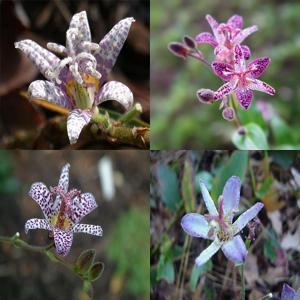} \\
\includegraphics[width=0.15\linewidth]{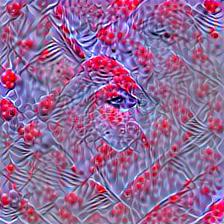} &
\includegraphics[width=0.15\linewidth]{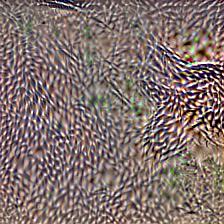} &
\includegraphics[width=0.15\linewidth]{fig/texture_images/cub/189_Red_bellied_Woodpecker.jpg} &
\includegraphics[width=0.15\linewidth]{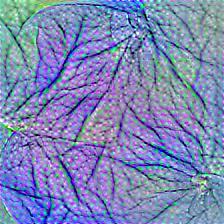} &
\includegraphics[width=0.15\linewidth]{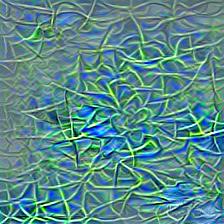} &
\includegraphics[width=0.15\linewidth]{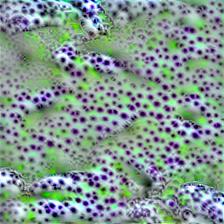} \\
\includegraphics[trim={1cm, 0, 0, 0}, clip, width=0.15\linewidth]{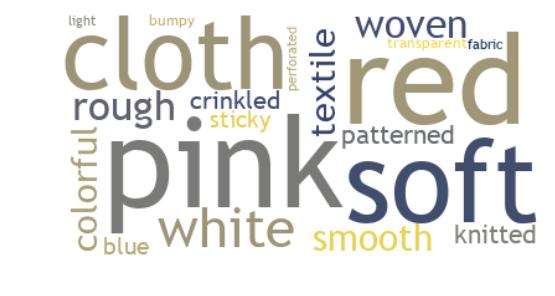} &
\includegraphics[trim={1cm, 0, 0, 0}, clip, width=0.15\linewidth]{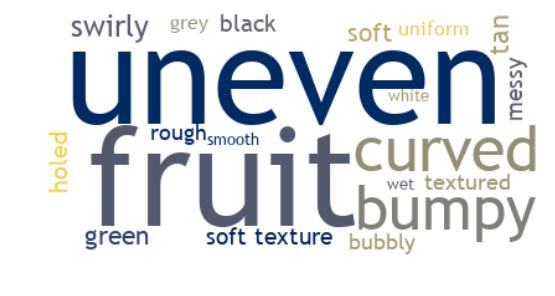} &
\includegraphics[trim={1cm, 0, 0, 0}, clip, width=0.15\linewidth]{fig/word_images/cub/189_Red_bellied_Woodpecker.jpg} &
\includegraphics[trim={1cm, 0, 0, 0}, clip, width=0.15\linewidth]{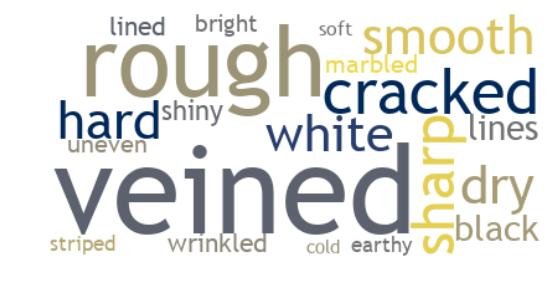} &
\includegraphics[trim={1cm, 0, 0, 0}, clip, width=0.15\linewidth]{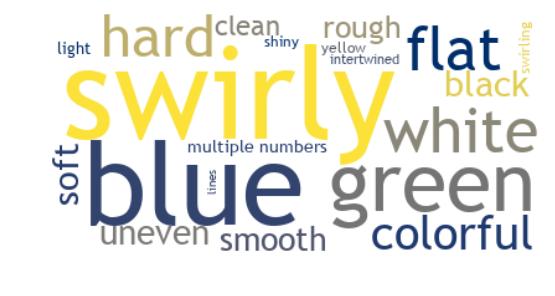} &
\includegraphics[trim={1cm, 0, 0, 0}, clip, width=0.15\linewidth]{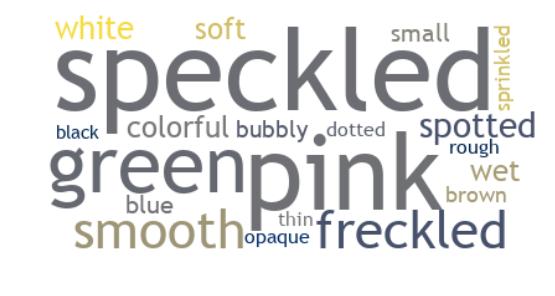} \\
\includegraphics[width=0.15\linewidth]{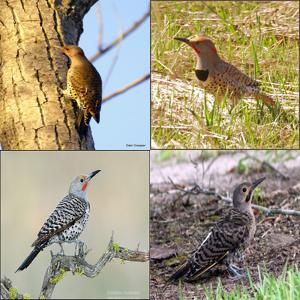} &
\includegraphics[width=0.15\linewidth]{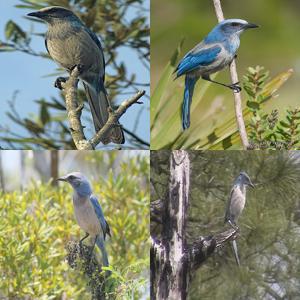} &
\includegraphics[width=0.15\linewidth]{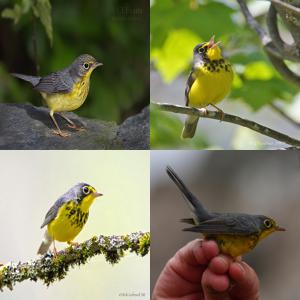} &
\includegraphics[width=0.15\linewidth]{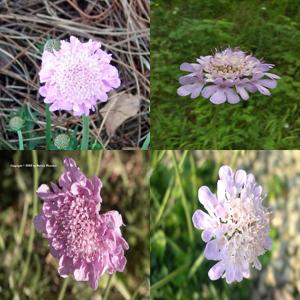} &
\includegraphics[width=0.15\linewidth]{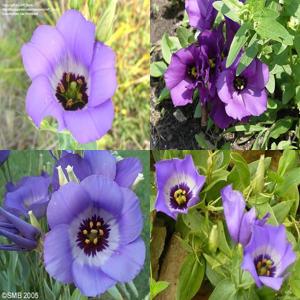} &
\includegraphics[width=0.15\linewidth]{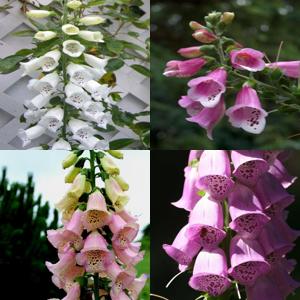} \\
\includegraphics[width=0.15\linewidth]{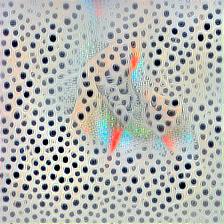} &
\includegraphics[width=0.15\linewidth]{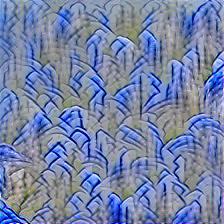} &
\includegraphics[width=0.15\linewidth]{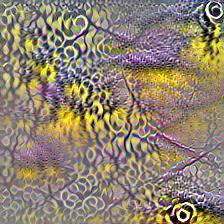} &
\includegraphics[width=0.15\linewidth]{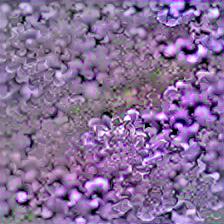} &
\includegraphics[width=0.15\linewidth]{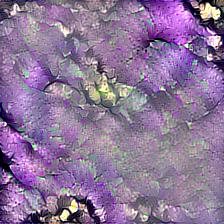} &
\includegraphics[width=0.15\linewidth]{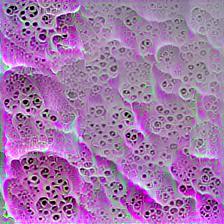} \\
\includegraphics[trim={1cm, 0, 0, 0}, clip, width=0.15\linewidth]{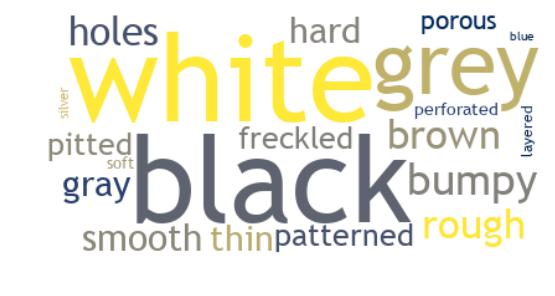} &
\includegraphics[trim={1cm, 0, 0, 0}, clip, width=0.15\linewidth]{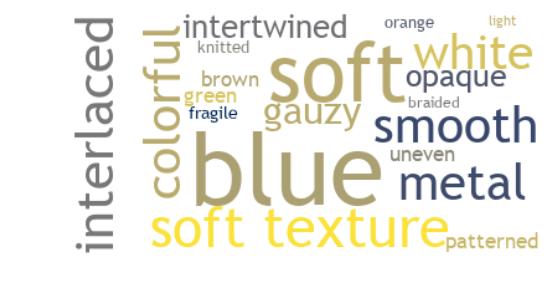} &
\includegraphics[trim={1cm, 0, 0, 0}, clip, width=0.15\linewidth]{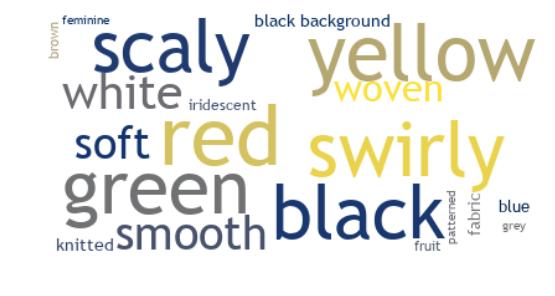} &
\includegraphics[trim={1cm, 0, 0, 0}, clip, width=0.15\linewidth]{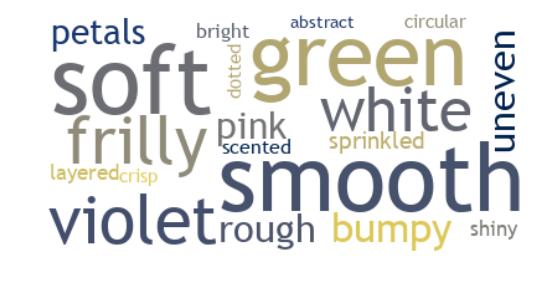} &
\includegraphics[trim={1cm, 0, 0, 0}, clip, width=0.15\linewidth]{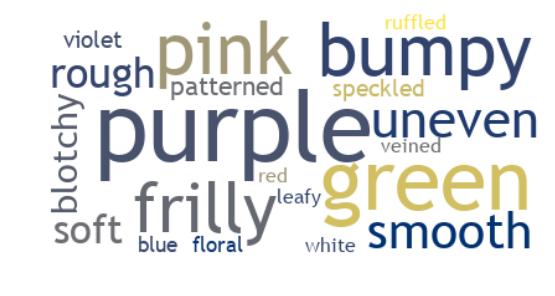} &
\includegraphics[trim={1cm, 0, 0, 0}, clip, width=0.15\linewidth]{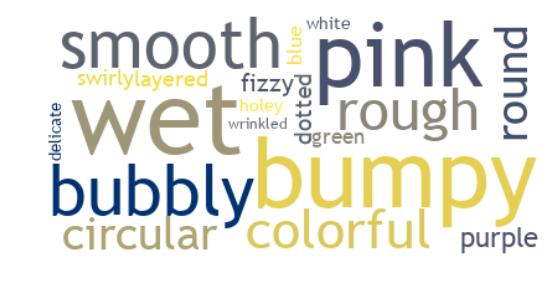} \\
\end{tabular}
\caption{\label{fig:viz1} Visualization of fine-grained categories from Caltech-UCSD birds and Oxford flowers. Each example is shown as a column of three images which consists of training examples (top), texture images (middle) and texture attributes as word clouds (bottom). The size of each phrase in the cloud reflects its likelihood of being associated with the maximal texture.}
\end{center}
\vspace{-1cm}
\end{figure*}

\begin{figure*}[t]
\begin{center}
\setlength{\tabcolsep}{2pt}
\begin{tabular}{cc|cc|cc}
\multicolumn{2}{c|}{\textbf{FGVC Butterflies and Moths}}  & \multicolumn{2}{c|}{\textbf{FGVC Fungi}} & \multicolumn{2}{c}{\textbf{FGVC Flowers}} \\
\includegraphics[width=0.15\linewidth]{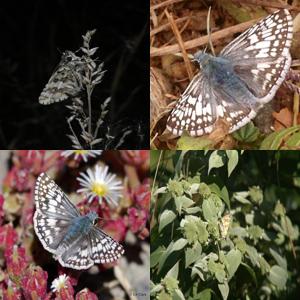} &
\includegraphics[width=0.15\linewidth]{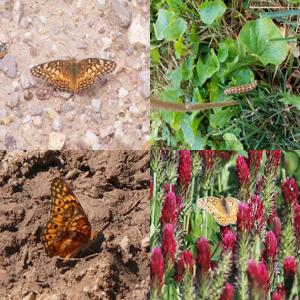} &
\includegraphics[width=0.15\linewidth]{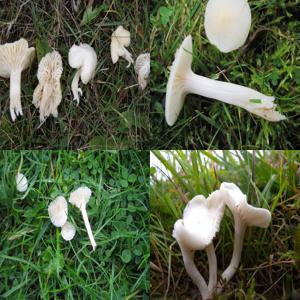} &
\includegraphics[width=0.15\linewidth]{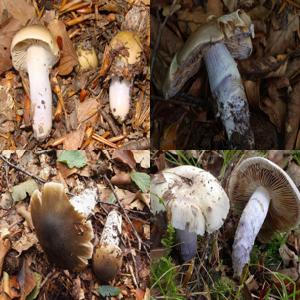} &
\includegraphics[width=0.15\linewidth]{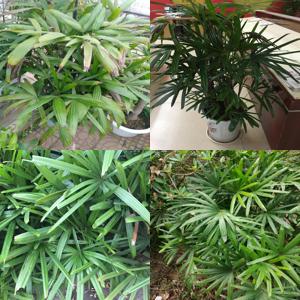} &
\includegraphics[width=0.15\linewidth]{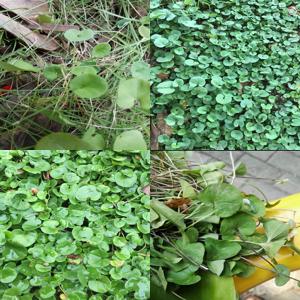} \\
\includegraphics[width=0.15\linewidth]{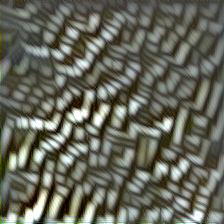} &
\includegraphics[width=0.15\linewidth]{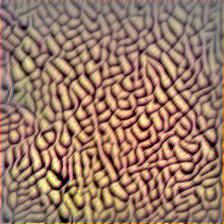} &
\includegraphics[width=0.15\linewidth]{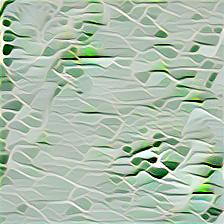} &
\includegraphics[width=0.15\linewidth]{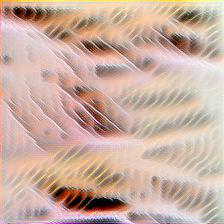} &
\includegraphics[width=0.15\linewidth]{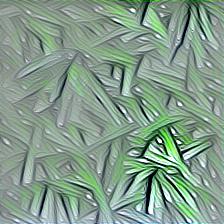} &
\includegraphics[width=0.15\linewidth]{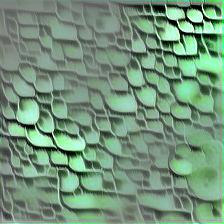} \\
\includegraphics[trim={1cm, 0, 0, 0}, clip, width=0.15\linewidth]{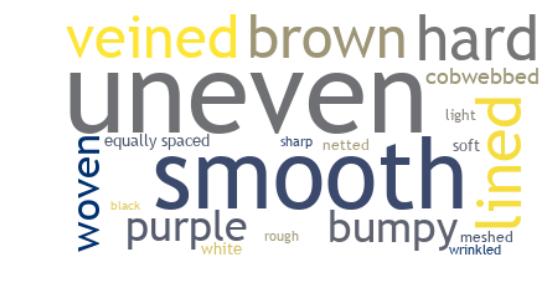} &
\includegraphics[trim={1cm, 0, 0, 0}, clip, width=0.15\linewidth]{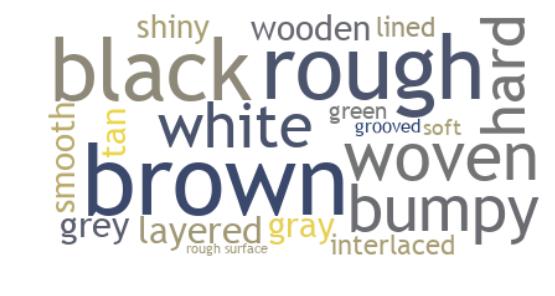} &
\includegraphics[trim={1cm, 0, 0, 0}, clip, width=0.15\linewidth]{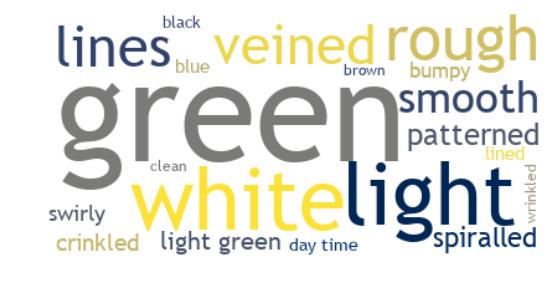} &
\includegraphics[trim={1cm, 0, 0, 0}, clip, width=0.15\linewidth]{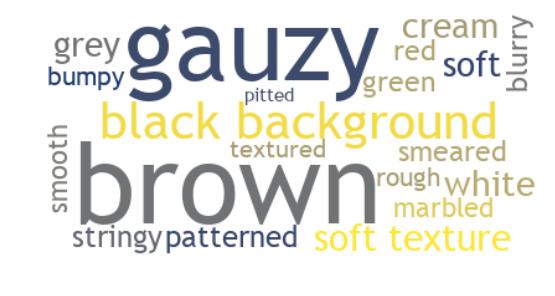} &
\includegraphics[trim={1cm, 0, 0, 0}, clip, width=0.15\linewidth]{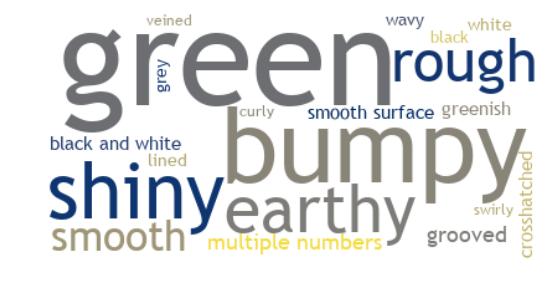} &
\includegraphics[trim={1cm, 0, 0, 0}, clip, width=0.15\linewidth]{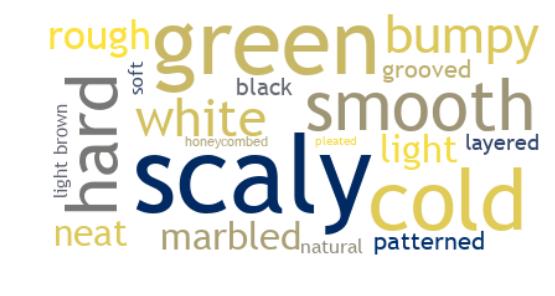} \\
\includegraphics[width=0.15\linewidth]{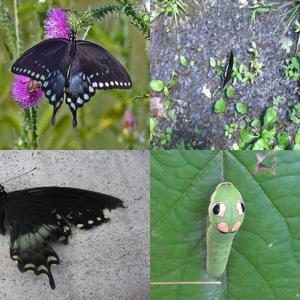} &
\includegraphics[width=0.15\linewidth]{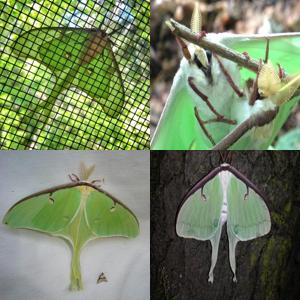} &
\includegraphics[width=0.15\linewidth]{fig/montage_images/fungi/Boletus_reticulatus.jpg} &
\includegraphics[width=0.15\linewidth]{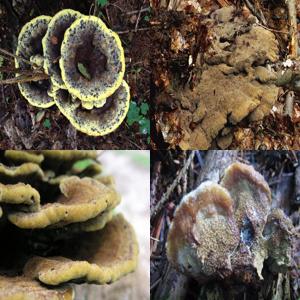} &
\includegraphics[width=0.15\linewidth]{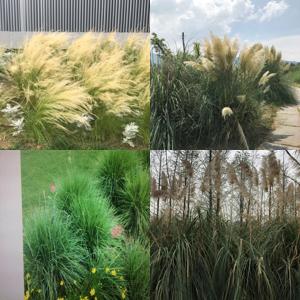} &
\includegraphics[width=0.15\linewidth]{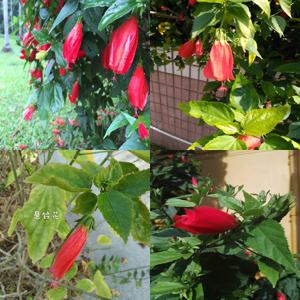} \\
\includegraphics[width=0.15\linewidth]{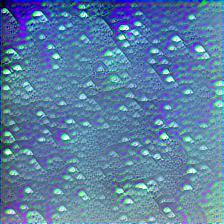} &
\includegraphics[width=0.15\linewidth]{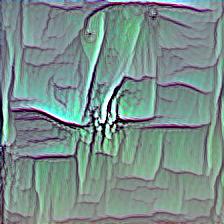} &
\includegraphics[width=0.15\linewidth]{fig/texture_images/fungi/Boletus_reticulatus.jpg} &
\includegraphics[width=0.15\linewidth]{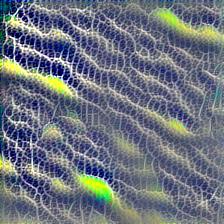} &
\includegraphics[width=0.15\linewidth]{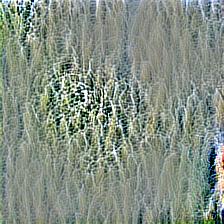} &
\includegraphics[width=0.15\linewidth]{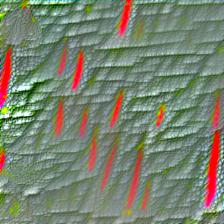} \\
\includegraphics[trim={1cm, 0, 0, 0}, clip, width=0.15\linewidth]{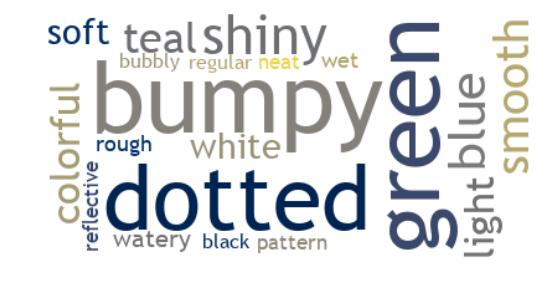} &
\includegraphics[trim={1cm, 0, 0, 0}, clip, width=0.15\linewidth]{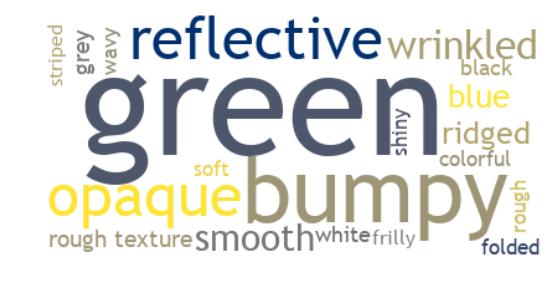} &
\includegraphics[trim={1cm, 0, 0, 0}, clip, width=0.15\linewidth]{fig/word_images/fungi/Boletus_reticulatus.jpg} &
\includegraphics[trim={1cm, 0, 0, 0}, clip, width=0.15\linewidth]{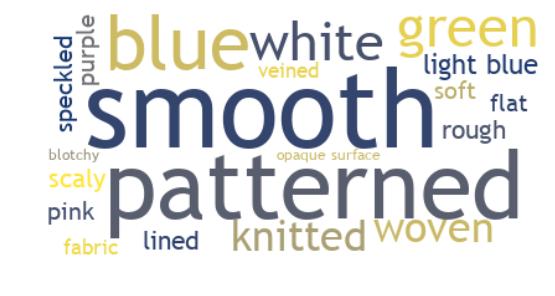} &
\includegraphics[trim={1cm, 0, 0, 0}, clip, width=0.15\linewidth]{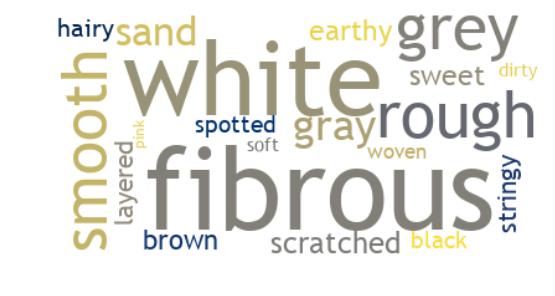} &
\includegraphics[trim={1cm, 0, 0, 0}, clip, width=0.15\linewidth]{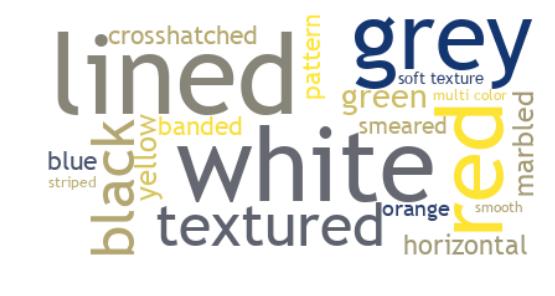} \\
\includegraphics[width=0.15\linewidth]{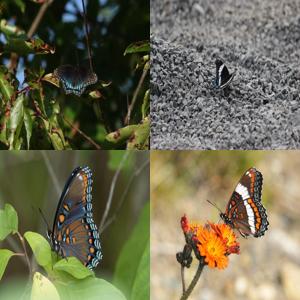} &
\includegraphics[width=0.15\linewidth]{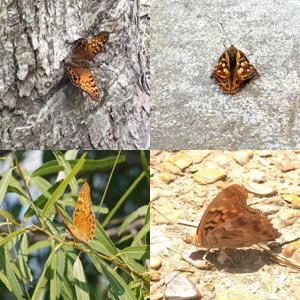} &
\includegraphics[width=0.15\linewidth]{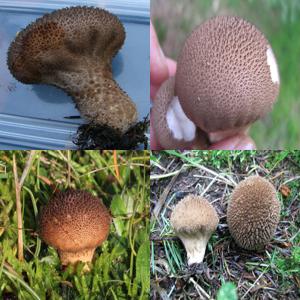} &
\includegraphics[width=0.15\linewidth]{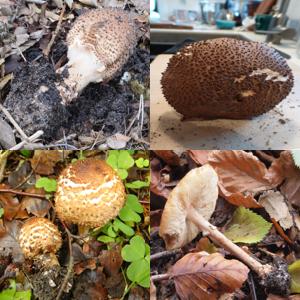} &
\includegraphics[width=0.15\linewidth]{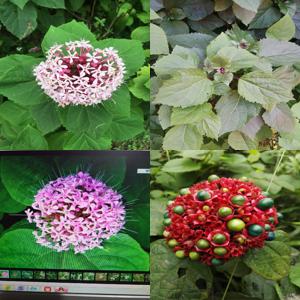} &
\includegraphics[width=0.15\linewidth]{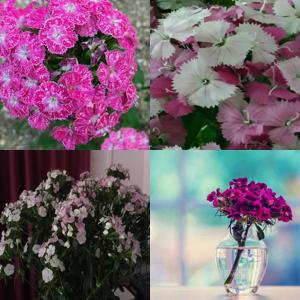} \\
\includegraphics[width=0.15\linewidth]{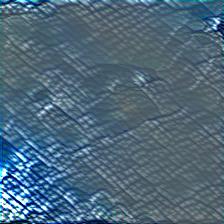} &
\includegraphics[width=0.15\linewidth]{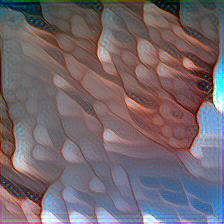} &
\includegraphics[width=0.15\linewidth]{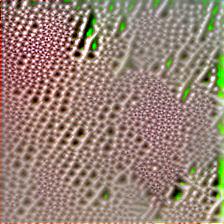} &
\includegraphics[width=0.15\linewidth]{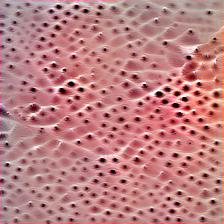} &
\includegraphics[width=0.15\linewidth]{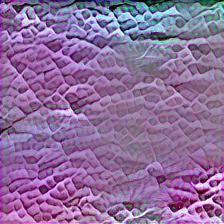} &
\includegraphics[width=0.15\linewidth]{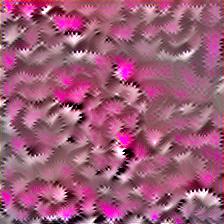} \\ 
\includegraphics[trim={1cm, 0, 0, 0}, clip, width=0.15\linewidth]{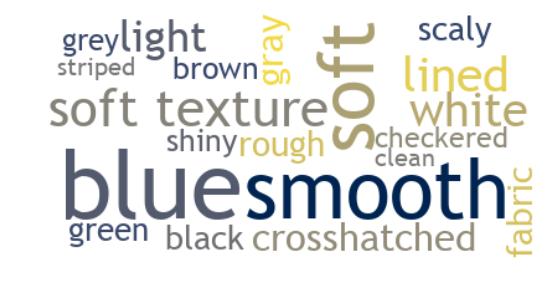} &
\includegraphics[trim={1cm, 0, 0, 0}, clip, width=0.15\linewidth]{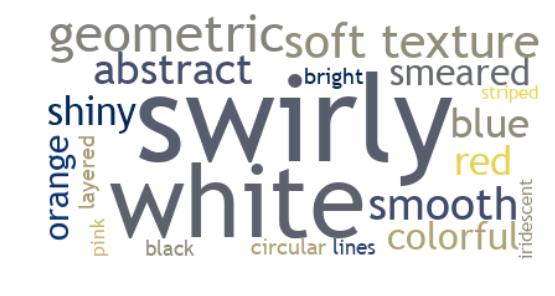} &
\includegraphics[trim={1cm, 0, 0, 0}, clip, width=0.15\linewidth]{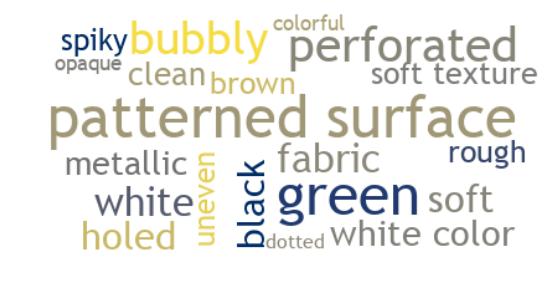} &
\includegraphics[trim={1cm, 0, 0, 0}, clip, width=0.15\linewidth]{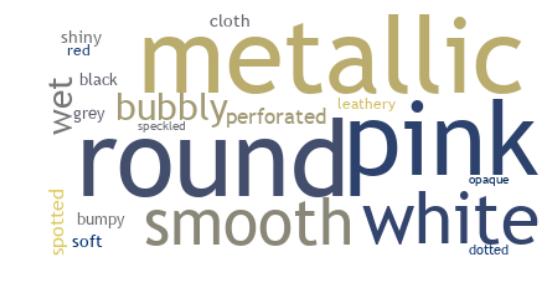} &
\includegraphics[trim={1cm, 0, 0, 0}, clip, width=0.15\linewidth]{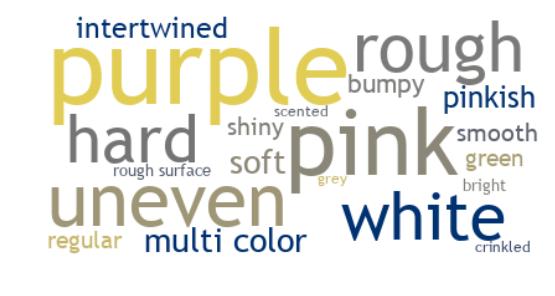} &
\includegraphics[trim={1cm, 0, 0, 0}, clip, width=0.15\linewidth]{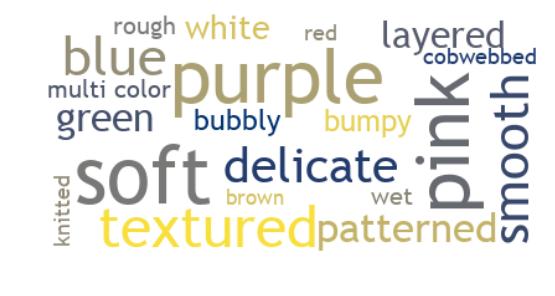} \\
\end{tabular}
\caption{\label{fig:viz2} Visualization of fine-grained categories from FGVC butterflies and moths, fungi, and flowers. Each example is shown as a column of three images which consists of training examples (top), texture images (middle) and texture attributes as word clouds (bottom). The size of each phrase in the cloud reflects its likelihood of being associated with the maximal texture.}
\end{center}
\vspace{-1cm}
\end{figure*}

Let $C_i$ be the predicted probability from layer $i$. Then the maximal inverse image for a target class $\hat{C}$ is
obtained as: $\min_\mathbf{x}\sum_{i=1}^{m} L\left(C_{i}, \hat{C}\right) + \gamma \Gamma(\mathbf{x}).$
Here $L$ is the softmax loss and $\Gamma(\mathbf{x})$ is the TV norm
that acts as a smoothness prior. This technique was also used to
visualize inverse images in~\cite{mahendran16visualizing}.
Figure~\ref{fig:splash} show the maximal images for three categories
along with their texture attributes. Additional visualizations selected arbitrarily across
datasets are shown in Figure~\ref{fig:viz1} and~\ref{fig:viz2}.
The maximal images indicate what discriminative texture properties are
learned from training images for classification of instances which often appear in clutter, with
wide ranges of pose and lighting variations, and under occlusions.

\vspace{-0.2in}
\paragraph{Describing maximal textures.} 
In addition, we provide the preliminary experiments on describing these
textures using attribute phrases that provide a language-based
explanation of discriminative texture properties.

We collected a new dataset with natural language descriptions of texture
details based on the Describable Textures Dataset
(DTD)~\cite{cimpoi14describing}. For each image from DTD, we ask five
human annotators to provide several attribute phrases (e.g., ``black and white dots'',
or ``colorful patterns''). 
We trained linear classifiers based on ResNet-101~\cite{he2016deep} activations to predict the probability of
each attribute phrase on our collected dataset.
For each maximal texture image, the ``phrase cloud'' shows the top 20 attribute phrases, with the font size proportional to the
predicted probability.

{\small
\bibliographystyle{ieee_fullname}
\bibliography{egbib}
}

\end{document}